\documentclass{article}

\usepackage[preprint]{corl_2026} 

\usepackage[utf8]{inputenc}
\usepackage{amsmath,amssymb}
\usepackage{booktabs,tabularx,array}
\usepackage[table,dvipsnames]{xcolor}
\usepackage{graphicx}
\usepackage{enumitem}
\usepackage{graphicx}
\usepackage{subcaption}
\usepackage[printonlyused]{acronym}
\DeclareUnicodeCharacter{2212}{\ensuremath{-}}

\usepackage{siunitx}
\newcommand{\parunit}[1]{\(\left(\unit{#1}\right)\)}

\newcommand{\vect}[1]{\mathbf{#1}}
\newcommand{\mat}[1]{\mathbf{#1}}
\newcommand{\dist}[1]{\left\|#1\right\|}

\DeclareMathOperator*{\median}{median}

\usepackage{url}
\usepackage{bookmark}
\usepackage[english,noabbrev,nameinlink,capitalise]{cleveref}
\newcommand\pcref[1]{(\cref{#1})}

\crefname{figure}{Fig.}{Figs.}
\Crefname{figure}{Figure}{Figures}

\definecolor{knn}{HTML}{EB6834}
\definecolor{learned}{HTML}{2A78D6}
\definecolor{tint}{HTML}{F5F6F8}
\newcommand{\knnword}{\textcolor{knn}{\textbf{\ac{kNN}}}}

\DeclareSIUnit{\pp}{\%pt}
\DeclareSIUnit{\pixel}{px}

\newcommand{\KnnAvg}{91.7}
\newcommand{\KnnPlainResnet}{88.7}
\newcommand{\KnnLearnt}{93.0}

\newcommand{\KnnPiZeroBaseLt}{86.9}
\newcommand{\KnnPiZeroLt}{17.9}

\newcommand{\PlusN}{420}
\newcommand{\PlusPerCat}{60}

\newcommand{\KnnUnseenStates}{91.6}

\newcommand{\FtN}{8}
\newcommand{\FtUp}{7}
\newcommand{\FtUpMin}{5}
\newcommand{\FtUpMax}{37}
\newcommand{\FtCleanUp}{5}
\newcommand{\SubtractMax}{48}

\newcommand{\FtDropBeforeMin}{20.9}
\newcommand{\FtDropBeforeMax}{34.7}
\newcommand{\FtDropAfterMin}{10.4}
\newcommand{\FtDropAfterMax}{21.0}

\newcommand{\LookupHuman}{0.78}
\newcommand{\LookupMin}{0.63}
\newcommand{\LookupMax}{0.93}
\newcommand{\LookupAbove}{5}
\newcommand{\LookupN}{8}
\newcommand{\LookupDistMin}{0.62}
\newcommand{\LookupDistMax}{0.99}

\newcommand{\NmOtherMin}{0.76}
\newcommand{\NmOtherMax}{0.88}
\newcommand{\NmFairRho}{0.61}
\newcommand{\NmFairR}{0.78}

\newcommand{\NmPiZeroLibero}{0.55}
\newcommand{\MapN}{4,150}
\newcommand{\MapGoalResnet}{79}
\newcommand{\MapGoalPromptMin}{96}
\newcommand{\MapGoalCausalMin}{76}
\newcommand{\MapGoalCausalMax}{82}
\newcommand{\MapPhaseN}{8}
\newcommand{\MapPhaseDown}{7}
\newcommand{\MapPhaseEvoBefore}{0.031}
\newcommand{\MapPhaseEvoAfter}{0.022}
\newcommand{\MapPhasePiZeroBefore}{0.028}
\newcommand{\MapPhasePiZeroAfter}{0.052}

\newcommand{\MapSuiteResnet}{85}
\newcommand{\MapSuiteBeforeMin}{40}
\newcommand{\MapSuiteBeforeMax}{84}
\newcommand{\MapSuiteAfterMin}{1}
\newcommand{\MapSuiteAfterMax}{43}
\newcommand{\MapSuiteDown}{8}
\newcommand{\MapTaskMin}{84}
\newcommand{\MapTaskMax}{100}
\newcommand{\MapPcaVarMin}{9}
\newcommand{\MapPcaVarMax}{24}
\newcommand{\OptunaN}{24}

\acrodef{VLA}{vision-language-action}
\acrodef{VLM}{vision-language model}
\acrodef{kNN}[\textit{k}-NN]{\textit{k}-nearest neighbors}
\acrodef{LIBERO}{LIfelong learning BEnchmark on RObot manipulation tasks}
\acrodef{LAC}{Local Action Consistency}
\acrodef{t-SNE}{t-distributed stochastic neighbor embedding}

\setlist{leftmargin=1.3em, itemsep=0.15em, topsep=0.3em, parsep=0pt}

\title{How Much of Your VLA is kNN}

\author{
  Denis Grachev\textsuperscript{1} \quad
  Damien LaRocque\textsuperscript{1} \quad
  Omar Moured\textsuperscript{1,2} \\
  \textsuperscript{1}Sereact GmbH, Stuttgart, Germany \\
  \textsuperscript{2}Karlsruher Institut für Technologie, Karlsruhe, Germany \\
  \texttt{\{denis.grachev,damien.larocque,omar.moured\}@sereact.ai}
}

\begin{document}
\maketitle

\begin{abstract}
    \Ac{VLA} models now exceed \SI{95}{\percent} success on the \acs{LIBERO} simulation benchmark, yet it remains unclear how much of this performance requires generalization beyond the training demonstrations.
    In this work, we show that retrieving nearby demonstrations and reusing their action chunks achieves a success rate of \SI{91.7}{\percent} on \acs{LIBERO} without policy training.
    %
    %
    Under LIBERO-Plus perturbations, retrieval shows the smallest drop of all evaluated policies.
    Across nine publicly available \acp{VLA}, fine-tuning mostly makes the vision-language backbone a better retrieval key, and the actions of the \acp{VLA} remain close to the demonstrations.
    %
    In addition, we introduce \ac{LAC}, an offline metric that measures whether nearby observations in representation space are also associated with similar future actions.
    We show that \ac{LAC} tracks the success rate of retrieval and can guide the training of a retrieval key.
    %
    Our evaluation and source code are available at \url{https://github.com/sereact/lerobot\_knn}.
\end{abstract}


\acresetall

\section{Introduction}\label{sec:introduction}

Modern \ac{VLA} models typically combine a large pretrained vision-language backbone with an action-generation head (e.g., regression, diffusion, or flow matching).
These models are further pretrained on large, heterogeneous collections of robot trajectories and subsequently fine-tuned for downstream manipulation tasks~\citep{pi2025pi05}.
Their performance is commonly evaluated on standardized benchmarks, which measure closed-loop task success across predefined tasks and initial conditions.

Among these, the \ac{LIBERO}~\citep{liu2023libero} is one of the most widely used simulation benchmarks, consisting of \num{130} language-conditioned Franka manipulation tasks across five suites: LIBERO-Spatial, LIBERO-Object, LIBERO-Goal, with \num{10} tasks each, and LIBERO-90 and LIBERO-Long, with \num{90} and \num{10} tasks, respectively.
Each task comes with \num{50} demonstrations.
\acp{VLA} are commonly fine-tuned on these demonstrations and evaluated in closed loop from a fixed set of benchmark initial states.

Recent progress has pushed LIBERO close to saturation, with several modern \acp{VLA} reporting success rates around or above \SI{97}{\percent}~\citep{Kim2025,pi2025pi05,sun2026vlajepa}, making further gains increasingly difficult to interpret as improved generalization.
Yet, recent studies show that policy performance can degrade substantially under changes in viewpoint, robot initialization, language, appearance, or object layout~\citep{fei2025liberoplus}.
These findings raise the question of how much benchmark success reflects generalization beyond the demonstrations, and how much can be achieved through demonstration retrieval alone.

\begin{figure}[htbp]
    \centering
    \includegraphics[width=1\linewidth]{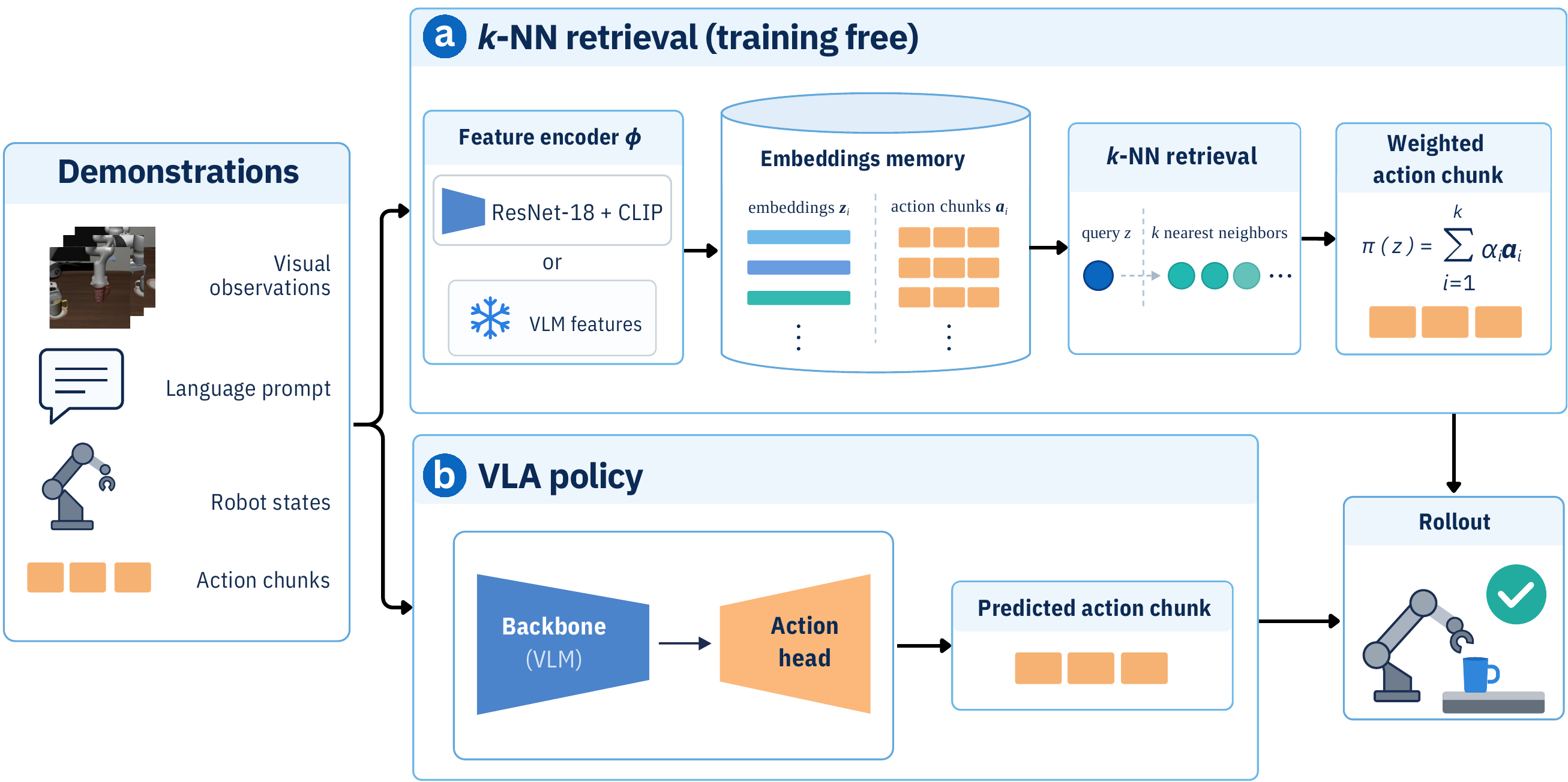}
    \caption{
        Overview of our evaluation pipeline.
        Both are evaluated with the same rollout protocol.
        (a) a \ac{kNN} retrieves neighboring demonstration embeddings and combines their associated action chunks;
        (b) the \ac{VLA} predicts an action chunk directly through its backbone and action head.
    }
    \label{fig:pipeline}
\end{figure}

To quantify how much benchmark performance can be explained by retrieval, we propose a training-free \ac{kNN} approach that retrieves demonstration states similar to the current observation and combines the action chunks that follow them, as shown in \cref{fig:pipeline}. Our contributions are as follows:

\begin{enumerate}
    \item Quantify the extent to which benchmark performance can be explained by retrieval alone.
    \item Use retrieval to evaluate the backbones of nine public \acp{VLA} before and after fine-tuning, and compare them with the full policies.
    \item Introduce \ac{LAC}, an offline metric of embedding quality for retrieval, and use it to guide the training of a retrieval key.
\end{enumerate}

\section{Related works}\label{sec:rw}

\paragraph{Retrieval in Robot Learning.}
Retrieval in robot learning has mainly been used for data selection, in-context adaptation, and non-parametric control.
For data selection, Behavior Retrieval~\citep{Du2023} retrieves task-relevant trajectories from large offline datasets, FlowRetrieval~\citep{lin2025flow} selects demonstrations using optical-flow-based motion similarity, and STRAP~\citep{memmel2025strap} retrieves reusable sub-trajectories from offline data to augment downstream policy training.
A second line of work uses retrieval at inference time: REGENT~\citep{sridhar2025regent} conditions a learned agent on retrieved experience, RICL~\citep{sridhar2025ricl} retrieves demonstrations as context for a pretrained VLA, and R+X~\citep{Papagiannis2025} retrieves relevant behaviors from human videos for in-context imitation.
In these approaches, retrieval primarily provides additional context to a learned policy, rather than producing actions.
Closer to our setting, VINN~\citep{Pari2022vinn} showed that nearest-neighbor regression over learned visual features can itself provide a competitive imitation policy, while \citet{he2025alt} showed that Diffusion Policy often generates action chunks close to training trajectories and proposed an explicit lookup-table alternative.
These works show that retrieval can either replace learned control or help explain its behavior.

Unlike most of these methods, our lookup requires no additional training.
It uses frozen, off-the-shelf encoders or the vision-language backbone of an existing \ac{VLA}, without updating their parameters on the benchmark.
This allows us to evaluate how well a simple retrieval-based method performs without learning an action policy.
When using the \ac{VLM} backbone of a \ac{VLA}, we can also compare retrieval-based action selection with the VLA's learned policy using the same underlying visual representations.

\paragraph{VLA Generalization and Memorization.}
Recent work has questioned whether strong in-distribution VLA performance translates to robust generalization.
LIBERO-Plus~\citep{fei2025liberoplus} shows that models exceeding \SI{95}{\percent} of success on standard LIBERO can degrade sharply under controlled changes in camera viewpoint, robot state, language, appearance, and layout, while LIBERO-PRO~\citep{zhou2025liberopro} further shows that some policies continue to execute memorized action patterns despite changes to objects or instructions.
A related line of work studies how VLA training affects generalization.
These studies indicate that supervised fine-tuning can remain brittle under distribution shift~\citep{Liu2025}, while OTTER~\citep{Huang25otter} and Knowledge Insulation~\citep{driess2025insulating} showed that adapting pretrained \acp{VLM} can overwrite useful semantics.
Our work complements these studies by directly measuring how action-relevant retrieval quality in the \ac{VLM} backbone changes before and after \ac{VLA} fine-tuning.

\section{Methodology}\label{sec:methodology}

We use a \ac{kNN} directly as a training-free policy.
Offline, each demonstration observation is embedded together with the action chunk that follows it.
At inference time, the current observation is embedded in the same space, its nearest demonstrated states are retrieved, and their associated action chunks are combined to produce the next action.

\subsection{Memory}

Let $\mathcal{D}=\{\tau_m\}_{m=1}^{M}$ denote the set of $M$ demonstration trajectories.
From these demonstrations, we construct a retrieval memory
\begin{equation}
    \mathcal{M}
    =
    \left\{
    \left(o_i,\vect{a}_{t_i:t_i+H},m_i\right)
    \right\}_{i=1}^{N},
    \label{eq:memory}
\end{equation}
where each entry contains a stored observation $o_i$, the action chunk of horizon $H$ that follows it, and the index $m_i$ of its source trajectory.
We store every second frame from the \num{1693} LIBERO demonstrations~\citep{liu2023libero}, resulting in approximately \(137\mathrm{k}\) memory entries.

\subsection{Retrieval Distance}

Given a query observation $q$ and memory entry $i$, we compute a weighted distance \(d(q,i)\) across the available modalities:
\begin{equation}
    d(q,i)
    =
    \sum_r
    \frac{w_r}{s_r}
    d_r(q,i),
    \label{eq:retrieval-distance}
\end{equation}
where $d_r$ is the distance for modality $r$, $w_r$ controls its contribution, and $s_r$ normalizes its scale.

The distance combines visual observations, robot state, language instruction, episode time, and recent observation history.
Visual and language terms are computed in their respective embedding spaces, while the exact representations and normalization are described in \cref{sec:implementation}.

\subsection{Neighbor selection and action prediction}\label{sec:methodology-neighbor-selection}

Let $\mathcal{N}_k(q)$ denote the $k$ memory entries with smallest retrieval distance to $q$.
To avoid overreliance on a single demonstration trajectory, we restrict the set to at most one entry per demonstration, such that
\begin{equation}
    m_i \neq m_j,
    \qquad
    \forall i \neq j,\ i,j \in \mathcal{N}_k(q).
    \label{eq:unique-demo}
\end{equation}

Each neighbor is associated with the action chunk that followed it in the demonstration.
We assign larger weights to closer neighbors using a softmax:
\begin{equation}
    \alpha_i
    =
    \frac{
        \exp\left(-d(q,i)/\tau\right)
    }{
        \sum_{j\in\mathcal{N}_k(q)}
        \exp\left(-d(q,j)/\tau\right)
    },
    \label{eq:knn-weights}
\end{equation}
where $\tau$ is the softmax temperature.

The predicted action chunk is then
\begin{equation}
    \pi(q)
    =
    \sum_{i\in\mathcal{N}_k(q)}
    \alpha_i
    \vect{a}_{t_i:t_i+H}.
    \label{eq:knn-policy}
\end{equation}

\subsection{\Acf{LAC}}\label{sec:lac}

\begin{figure}[htbp]
    \centering
    \includegraphics[width=1\linewidth]{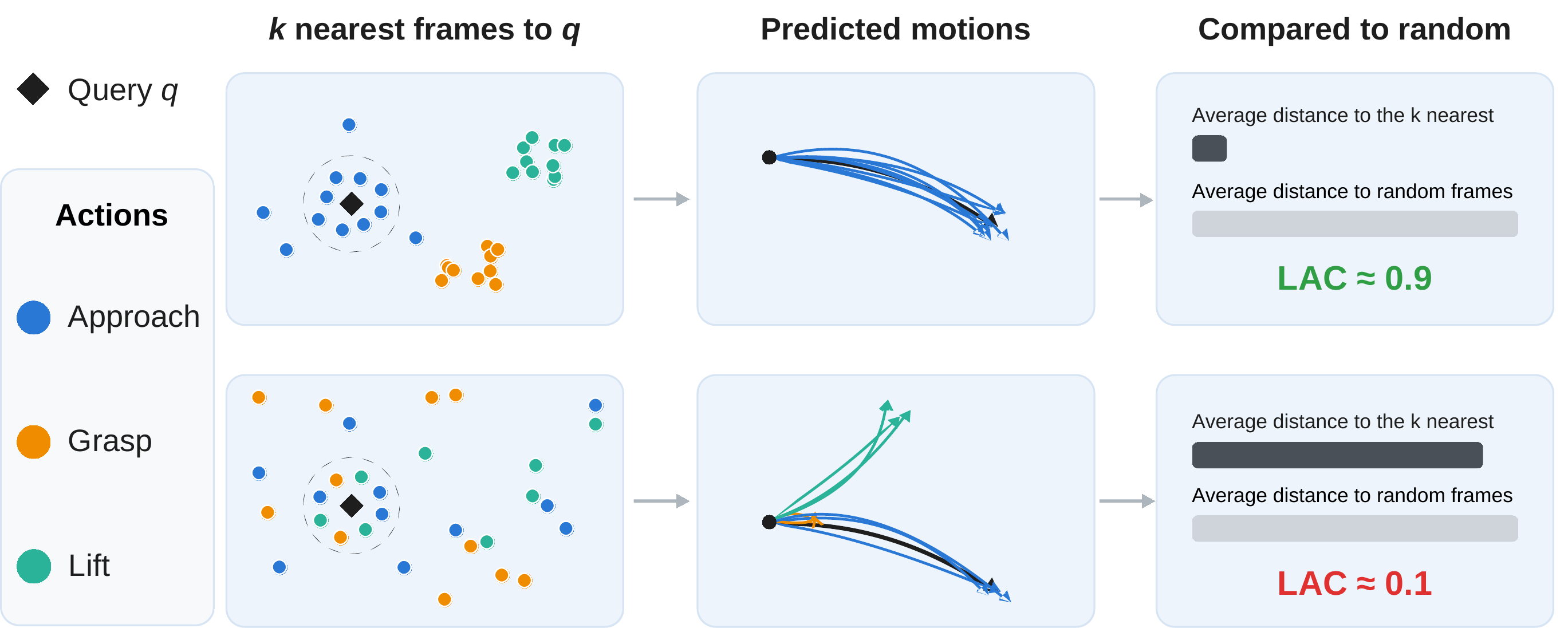}
    \caption{\Acf{LAC} is the ratio between the average action distance of the retrieved $k$ nearest neighbors and the average action distance over the same-task set.}
    \label{fig:lac}
\end{figure}

We propose \acf{LAC}, an offline metric for evaluating retrieval representations without rollouts, as illustrated in \cref{fig:lac}. For each held-out query \(q\), \ac{LAC} measures whether its nearest neighbors in representation space also exhibit similar future action chunks, relative to other demonstrations from the same task, such that
\begin{equation}
    \mathrm{LAC@}k(q)
    =
    1-
    \frac{
        \dfrac{1}{k}
        \sum\limits_{i\in\mathcal{N}_k(q)}
        \dist{\vect{a}_q-\vect{a}_i}_2^2
    }{
        \dfrac{1}{|\mathcal{C}(q)|}
        \sum\limits_{j\in\mathcal{C}(q)}
        \dist{\vect{a}_q-\vect{a}_j}_2^2
    },
    \label{eq:lac}
\end{equation}
where $\mathcal{N}_k(q)$ contains the $k$ nearest neighbors and $\mathcal{C}(q)$ contains all memory entries from the same task as the query $q$.
We report $\mathrm{LAC@}8$ averaged over held-out queries. Higher LAC indicates that nearby representations also correspond to similar future actions relative to typical same-task matches.

\section{Experimental Setup}\label{sec:experiments}

For \ac{LIBERO}~\citep{liu2023libero}, we evaluate on the Spatial, Object, Goal, and Long suites, comprising 40 tasks, and use \num{1693} training demonstrations.
Following the official protocol, we run 50 episodes per task, yielding \num{2000} rollouts per policy.
To assess robustness, we use LIBERO-Plus~\citep{fei2025liberoplus}, which introduces seven categories of perturbations to the same tasks.
We select a fixed subset of \PlusN{} variants, with \PlusPerCat{} variants per category, shared across all policies and held out from training and hyperparameter tuning.
We evaluate nine publicly available LIBERO VLA checkpoints from LeRobot~\citep{LeRobot2026}, using each model as released with its corresponding observation, action, and inference conventions.
Model-specific settings are provided in \cref{app:vlas}.

\section{Implementation Details}\label{sec:implementation}

\paragraph{kNN representation.}
We encode each camera view using a frozen ImageNet-pretrained ResNet-18~\citep{he2016resnet}, extracting a $4\times4$ spatial feature grid from layer 3. Robot proprioception is standardized using statistics from the demonstration memory, while language instructions are represented by unit-normalized CLIP embeddings~\citep{radford2021clip}. Retrieval combines distances over visual features, robot state, language, episode time, and observation history, following \cref{eq:retrieval-distance}.
Each term is normalized by its median within-task distance so that modalities with different numerical scales remain comparable.

\paragraph{Hyperparameters.}
We tune the \ac{kNN} hyperparameters with Optuna~\citep{akiba2019optuna}.
The complete search space, normalization variants, history terms, and ablations are reported in \cref{app:knn}.

\paragraph{Policy variants.}
We evaluate three variants of the \ac{kNN} policy, which differ in the retrieval distance, the visual representation, and the execution of action chunks.
In all variants, executed actions are weighted combinations of demonstrated actions.
\emph{k-NN (ResNet)} uses only the image distance, restricts retrieval to the demonstrations of the instructed task, and executes the full weighted action chunk of its $k=8$ nearest entries before retrieving again.
\emph{k-NN (multimodal)} retrieves from all demonstrations with the full distance of \cref{eq:retrieval-distance}, which combines image features, robot state, language, episode time, and observation history, with $k=13$.
It additionally applies mode seeking, which down-weights neighbors outside the dominant action mode, and temporal ensembling~\citep{zhao2023act}, which re-plans at every control step and combines the overlapping predictions.
\emph{k-NN (LAC key)} uses the same retrieval and execution, but applies a learned linear projection to the standardized ResNet features of each grid cell, trained with a differentiable surrogate of \ac{LAC} \pcref{eq:lac}: the chunk of a frame is predicted from the softmax-weighted chunks of its neighbors in other demonstrations.

\paragraph{VLA retrieval features.}
To probe the representations of each \ac{VLA}, we run the \emph{ResNet} variant with the hidden states of the image tokens at one \ac{VLM} layer in place of the ResNet-18 features.
These states are pooled to a $2\times2$ grid per camera and standardized per dimension.
The layer maximizes the $R^2$ of the lookup's action predictions on held-out demonstrations, averaged over the backbone before and after fine-tuning, so that both checkpoints use the same layer.

\section{Results}\label{sec:results}

We use retrieval to measure how much of \ac{VLA} performance can be recovered directly from demonstrations, and how closely \acp{VLA} remain tied to them.
We first compare \ac{kNN} policies with nine public \acp{VLA} on \ac{LIBERO} and under the perturbations of LIBERO-Plus, and identify the components that drive retrieval performance.
We then examine how \ac{LIBERO} fine-tuning changes the \ac{VLM} backbone as a retrieval key, and how closely the actions of the \acp{VLA} follow the demonstrations.
Finally, we evaluate whether \ac{LAC} can estimate the quality of a retrieval key without rollouts.

\subsection{Benchmark Results}\label{sec:benchmark-performance}

\Cref{tab:results} compares the success rates of the three \ac{kNN} variants defined in \cref{sec:implementation} with nine \acf{VLA} models on \ac{LIBERO} and LIBERO-Plus.
On average, the \emph{ResNet} variant achieves a success rate of \SI{88.7}{\percent} on \ac{LIBERO} and \SI{75.5}{\percent} on LIBERO-Plus.
The \emph{multimodal} variant reaches \SI{91.7}{\percent} and \SI{82.4}{\percent}, respectively, while the \emph{\ac{LAC} key} variant reaches \SI{93.0}{\percent} and \SI{81.7}{\percent}.
Although $\pi_{0.5}$ achieves the highest success rate on both benchmarks, the \ac{kNN} policies perform on par with $\pi_{0}$-FAST on \ac{LIBERO} and outperform SmolVLA and $\pi_{0}$.
These results show that retrieval from demonstrations alone can recover a large fraction of benchmark performance, without learning a control policy.

\begin{table}[t!]
\centering
\caption{Success rate (\unit{\percent}) on LIBERO (official protocol, \num{2000} episodes) and on the same 420 LIBERO-Plus variants. \acp{VLA} are ordered by average on LIBERO-Plus. $\Delta$ is the difference between the average success rates on LIBERO-Plus and LIBERO, computed before rounding. The \ac{kNN} rows are training-free, except for the \ac{LAC} key.}\label{tab:results}
\scriptsize
\setlength{\tabcolsep}{5pt}
\resizebox{\linewidth}{!}{%
  \begin{tabular}{@{}l*{5}{r}@{\hspace{1.5mm}}*{5}{r}@{\hspace{1.5mm}}r@{}}
  \toprule
  & \multicolumn{5}{c}{LIBERO}
  & \multicolumn{5}{c}{LIBERO-Plus}
  & \\
  \cmidrule(lr){2-6}\cmidrule(lr){7-11}
  & avg & Spat & Obj & Goal & Long
  & avg & Spat & Obj & Goal & Long & $\Delta$ \\
  \midrule

  $\pi_{0.5}$~\citep{pi2025pi05}
  & \textbf{97.7} & 97.0 & 99.8 & \textbf{97.8} & \textbf{96.2}
  & \textbf{85.5} & \textbf{90.5} & 83.8 & 82.9 & \textbf{84.8} & -12.2 \\

  MolmoAct2~\citep{fang2026molmoact2}
  & 97.5 & \textbf{97.4} & 99.4 & 97.6 & 95.6
  & 82.6 & 89.5 & 81.9 & 76.2 & 82.9 & -14.9 \\

  GR00T N1.7~\citep{nvidia2025gr00t}
  & 94.9 & 95.8 & 98.2 & 94.6 & 90.8
  & 80.2 & 89.5 & 74.3 & 79.0 & 78.1 & -14.6 \\

  VLA-JEPA~\citep{sun2026vlajepa}
  & 97.2 & 95.4 & \textbf{100.0} & 97.4 & 95.8
  & 74.0 & 81.0 & 70.5 & 77.1 & 67.6 & -23.1 \\

  Evo-1~\citep{lin2026evo1}
  & 95.4 & 92.0 & 98.6 & 97.4 & 93.6
  & 71.4 & 74.3 & 73.3 & 69.5 & 68.6 & -24.0 \\

  X-VLA~\citep{zheng2026xvla}
  & 94.5 & 94.6 & 98.6 & 92.6 & 92.2
  & 64.5 & 71.4 & 64.8 & 64.8 & 57.1 & -30.0 \\

  $\pi_0$~\citep{black2024pi0}
  & 71.9 & 70.0 & 86.4 & 85.8 & 45.2
  & 42.9 & 54.3 & 41.9 & 46.7 & 28.6 & -29.0 \\

  SmolVLA~\citep{shukor2025smolvla}
  & 82.2 & 78.8 & 92.2 & 88.0 & 69.6
  & 27.9 & 30.5 & 49.5 & 11.4 & 20.0 & -54.3 \\

  $\pi_0$-FAST~\citep{pertsch2025fast}
  & 90.6 & 90.4 & 93.8 & 90.6 & 87.4
  & 25.7 & 39.0 & 34.3 & 16.2 & 13.3 & -64.8 \\

  \midrule

  \rowcolor{tint}\knnword{} (\emph{ResNet})
  & 88.7 & 90.4 & 95.0 & 95.2 & 74.2
  & 75.5 & 77.1 & 81.9 & 79.0 & 63.8 & -13.2 \\

  \rowcolor{tint}\knnword{} (\emph{multimodal})
  & 91.7 & 88.0 & 99.8 & 97.6 & 81.4
  & 82.4 & 79.0 & \textbf{85.7} & \textbf{86.7} & 78.1 & \textbf{-9.3} \\

  \rowcolor{tint}\knnword{} (\emph{LAC key})
  & 93.0 & 90.6 & 99.6 & 97.0 & 84.8
  & 81.7 & 87.6 & 83.8 & 81.0 & 74.3 & -11.3 \\

  \bottomrule
  \end{tabular}%
}
\end{table}

VLA-JEPA, Evo-1, and X-VLA exceed \SI{94}{\percent} on \ac{LIBERO}, but their success rates on LIBERO-Plus fall below that of the training-free \emph{ResNet} variant.
Under perturbation, the \emph{multimodal} \ac{kNN} outperforms 7 of the 9 \acp{VLA} and shows the smallest performance drop of all evaluated policies (\SI{9.3}{\pp}).
These results highlight that strong performance on \ac{LIBERO} does not necessarily translate to robustness under the perturbations of LIBERO-Plus.

\subsection{Ablation study}\label{sec:ablations}

To identify which components drive retrieval performance, we ablate the \emph{multimodal} \ac{kNN} policy one component at a time, as shown by \cref{tab:ablations} in \cref{app:ablations}.
Using only \SI{10}{\percent} of the demonstrations reduces success from \SI{91.7}{\percent} to \SI{77.4}{\percent}, and using raw pixels instead of embeddings lowers it to \SI{57.2}{\percent}.
Ignoring the language prompt causes a drop of \SI{27.1}{\pp}, whereas removing mode seeking, temporal ensembling, history, robot state, or the time step changes success by at most \SI{1}{\pp}, within the \SI{95}{\percent} confidence interval of $\pm\SI{1.2}{\pp}$ over \num{2000} episodes.
For $k$ between 4 and 32, success remains between \SI{89.4}{\percent} and \SI{92.7}{\percent}, and drops to \SI{78.6}{\percent} when only the nearest demonstration is replayed ($k=1$).
Overall, the visual representation, the language prompt, and the number of demonstrations account for most of the performance, while the remaining refinements contribute less.
Full ablations and hyperparameter studies are reported in \cref{app:knn}.

\subsection{Representation Analysis}\label{sec:representation-analysis}

\Cref{tab:backbones} shows that \ac{LIBERO} fine-tuning generally makes the \ac{VLM} backbone a better retrieval key, mostly under perturbation.
Retrieval with the fine-tuned backbone improves for \FtCleanUp{} of the \FtN{} fine-tuned \acp{VLA} on \ac{LIBERO}, and for \FtUp{} of \FtN{} on LIBERO-Plus, by \SIrange[range-units = single]{\FtUpMin}{\FtUpMax}{\pp}.
For these \acp{VLA}, the drop from \ac{LIBERO} to LIBERO-Plus also decreases, from \SIrange[range-units = single, range-phrase = --]{\FtDropBeforeMin}{\FtDropBeforeMax}{\pp} before fine-tuning to \SIrange[range-units = single, range-phrase = --]{\FtDropAfterMin}{\FtDropAfterMax}{\pp} after.
However, the effect is not universal: after fine-tuning, retrieval with the $\pi_0$ backbone drops from \SI{\KnnPiZeroBaseLt}{\percent} to \SI{\KnnPiZeroLt}{\percent} on \ac{LIBERO}.

\begin{table}[t!]
\centering
\caption{Lookup success using each \ac{VLA}'s \ac{VLM} backbone before and after \ac{LIBERO} fine-tuning, compared with the full \ac{VLA}. \textbf{Bold} indicates cases where retrieval on the fine-tuned backbone outperforms the corresponding VLA. $^{*}$GR00T freezes its \ac{VLM}, so before $=$ after.}\label{tab:backbones}
\footnotesize\setlength{\tabcolsep}{3pt}
\begin{tabularx}{\linewidth}{@{}l*{6}{>{\raggedleft\arraybackslash}X}@{}}
\toprule
& \multicolumn{3}{c}{LIBERO} & \multicolumn{3}{c}{LIBERO-Plus} \\
\cmidrule(lr){2-4}\cmidrule(lr){5-7}
& \knnword{} before & \knnword{} after & VLA & \knnword{} before & \knnword{} after & VLA \\
\midrule
$\pi_{0.5}$ & 85.4 & 81.8 & 97.7 & 60.5 & 65.7 & 85.5 \\
MolmoAct2 & 78.6 & 91.4 & 97.5 & 52.6 & 81.0 & 82.6 \\
GR00T N1.7* & 87.6 & 87.6 & 94.9 & 66.4 & 66.4 & 80.2 \\
VLA-JEPA & 84.0 & 83.0 & 97.2 & 49.3 & 65.7 & 74.0 \\
Evo-1 & 62.1 & 87.1 & 95.4 & 35.5 & \textbf{72.6} & 71.4 \\
X-VLA & 83.3 & 88.6 & 94.5 & 62.4 & \textbf{78.1} & 64.5 \\
$\pi_0$ & 86.9 & 17.9 & 71.9 & 56.7 & 2.4 & 42.9 \\
SmolVLA & 76.3 & \textbf{86.5} & 82.2 & 45.0 & \textbf{69.3} & 27.9 \\
$\pi_0$-FAST & 86.6 & \textbf{95.0} & 90.6 & 58.3 & \textbf{74.0} & 25.7 \\
\bottomrule
\end{tabularx}
\end{table}

For some models, retrieval on the fine-tuned backbone alone outperforms the full \ac{VLA}: for 2 of 9 models on \ac{LIBERO} and 4 of 9 on LIBERO-Plus \pcref{tab:backbones}, by up to \SI{\SubtractMax}{\pp}.
For these models, the results suggest that the action head does not exploit all the action-relevant information encoded in the backbone.

\begin{figure}[htbp]
    \centering
    \includegraphics[width=0.95\linewidth]{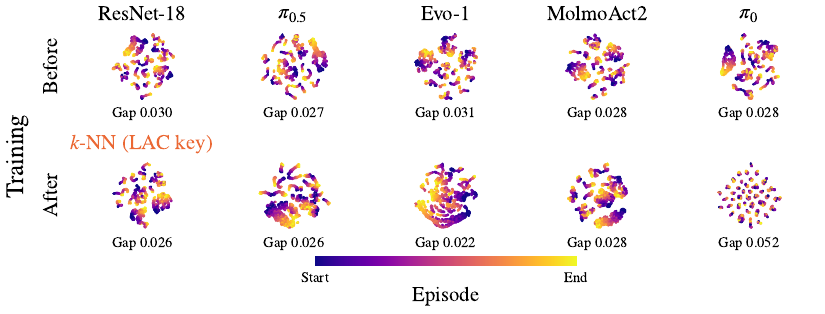}
    \caption{
        \ac{t-SNE} maps of retrieval keys for the same \MapN{} demonstration frames from \ac{LIBERO}.
        Top: before \ac{LIBERO} fine-tuning (ImageNet ResNet-18 for the first column).
        Bottom: after fine-tuning (\ac{LAC}-trained key for the first column).
        Colors indicate normalized episode progress.
        Below each map: median progress gap to the nearest frame from another demonstration of the same task.
    }
    \label{fig:maps}
\end{figure}

\paragraph{Changes in Local Representation Structure.}
To understand how fine-tuning affects the geometry of the representations, \cref{fig:maps} shows \ac{t-SNE} maps~\citep{vanderMaaten2008} of the retrieval keys of four \acp{VLA} before and after fine-tuning, and of the ImageNet ResNet-18 before and after \ac{LAC} training.
After fine-tuning, the keys of Evo-1 form trajectories that follow task progress, and the median progress gap between a frame and its nearest same-task neighbor from another demonstration decreases from \num{\MapPhaseEvoBefore} to \num{\MapPhaseEvoAfter}.
Across all backbones \pcref{fig:maps-tsne}, this gap decreases for \MapPhaseDown{} of the \MapPhaseN{} fine-tuned \acp{VLA}, but the change is small for $\pi_{0.5}$ and MolmoAct2.
$\pi_0$ is different: its fine-tuned key collapses into one cluster per task, and its nearest neighbors remain of the same task, but its progress gap increases from \num{\MapPhasePiZeroBefore} to \num{\MapPhasePiZeroAfter}, consistent with the collapse of its retrieval performance.
After fine-tuning, the main directions of variation of the keys also depend less on the \ac{LIBERO} suite, and thus on the scene, for all \MapSuiteDown{} backbones \pcref{fig:maps-pca}.

\subsection{Action-Space Analysis}\label{sec:action-space-analysis}

\begin{figure}[htbp]
    \centering
    \begin{subfigure}[t]{0.48\linewidth}
        \centering
        \includegraphics[width=\linewidth]{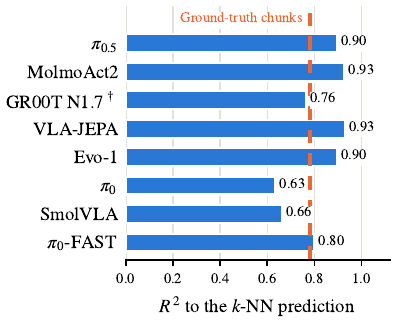}
        \caption{Agreement with the \ac{kNN} prediction.}
        \label{subfig:vla_lookup_r2}
    \end{subfigure}
    \hfill
    \begin{subfigure}[t]{0.45\linewidth}
        \centering
        \includegraphics[width=\linewidth]{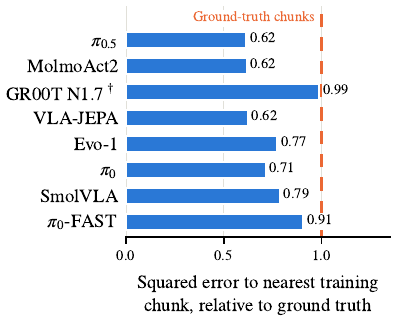}
        \caption{Squared error to the nearest training chunk.}
        \label{subfig:vla_lookup_distance}
    \end{subfigure}
    \caption{
        Comparison of the action chunks of each \ac{VLA}, in its \ac{LIBERO} rollouts, with the training demonstrations.
        (\subref{subfig:vla_lookup_r2}) $R^2$ between the \ac{VLA} chunks and the prediction of the \emph{ResNet} \ac{kNN} variant; higher indicates closer agreement.
        (\subref{subfig:vla_lookup_distance}) Median squared error to the nearest training chunk of the same task, relative to held-out ground-truth chunks; lower indicates greater proximity to the training data.
        Dashed lines: held-out ground-truth chunks.
        $^\dagger$Evaluated on LIBERO-Spatial only.
        X-VLA is excluded because it predicts absolute poses.
    }
    \label{fig:lookup}
\end{figure}

We next examine how closely the actions produced by the \acp{VLA} follow the demonstrations.
At each state visited in the \ac{LIBERO} rollouts of a \ac{VLA}, we compare its action chunk with the prediction of the \emph{ResNet} \ac{kNN} variant, the weighted mean of the chunks of its $k=8$ nearest demonstration frames of the same task.
As a reference, the same prediction matches held-out ground-truth chunks with $R^2=\num{\LookupHuman}$.
For the \acp{VLA}, $R^2$ ranges from \num{\LookupMin} to \num{\LookupMax} \pcref{subfig:vla_lookup_r2}: the chunks of \LookupAbove{} of the \LookupN{} \acp{VLA} are closer to the \ac{kNN} prediction than the ground-truth chunks are.
In addition, the chunks of all \acp{VLA} lie closer to a training chunk of the same task than held-out ground-truth chunks do, with a median squared error of \num{\LookupDistMin}-\num{\LookupDistMax} times the ground-truth reference \pcref{subfig:vla_lookup_distance}.
Together, these results suggest that, on \ac{LIBERO}, the actions of the \acp{VLA} remain close to the demonstrated actions.

\subsection{\ac{LAC} as an Offline Diagnostic}\label{sec:results-lac}

The previous analyses evaluate retrieval representations through closed-loop execution; we next ask whether \ac{LAC} \pcref{sec:lac} can estimate their quality from the demonstrations alone.
Excluding the fine-tuned \ac{VLM} backbones, which were trained on the held-out demonstrations, \ac{LAC} tracks closed-loop retrieval success (Pearson $r=\num{\NmFairR}$) but does not reliably rank keys of similar quality (Spearman $\rho=\num{\NmFairRho}$, \cref{fig:neighbors} in \cref{app:lac}).
It also flags the degraded fine-tuned $\pi_0$ backbone, with an \ac{LAC} of \num{\NmPiZeroLibero} against \num{\NmOtherMin}-\num{\NmOtherMax} for all other keys.
A linear projection of the ResNet features, trained with a differentiable surrogate of \ac{LAC}, raises the success rate of a plain \ac{kNN} lookup from \SI{\KnnPlainResnet}{\percent} to \SI{\KnnLearnt}{\percent}.

\section{Conclusion}\label{sec:conclusion}

In this work, we investigated how much of the performance observed on the \acf{LIBERO} can be recovered directly from demonstrations, without learning a control policy.
To this end, we used retrieval-based policies not only as baselines, but also as tools to analyze the representations and actions of existing \acp{VLA}.
First, retrieval alone recovers a substantial fraction of benchmark performance: a simple \ac{kNN} policy is competitive with several \acp{VLA} on \ac{LIBERO}, and its multimodal variant shows the smallest drop of all evaluated policies under the perturbations of LIBERO-Plus.
The actions of the \acp{VLA} also remain close to the demonstrations, suggesting that, on dense benchmarks such as \ac{LIBERO}, a large part of successful behavior can be recovered by matching test states to demonstrated ones and reusing their actions.
Second, retrieval allowed us to separate the contributions of the \ac{VLM} backbone and of the action head of a \ac{VLA}.
Across nine \acp{VLA}, the backbone already contains action-relevant structure, and fine-tuning mostly makes it a more robust retrieval key, improving retrieval under the perturbations of LIBERO-Plus for \FtUp{} of \FtN{} models.
This effect is not universal, as some representations degrade after fine-tuning.
For some models, retrieval from the fine-tuned backbone alone outperforms the full policy, suggesting that their action head does not exploit all the action-relevant information in the representation.
Third, we introduced \ac{LAC}, an offline metric that tracks closed-loop retrieval success, although it does not reliably rank representations of similar quality.
It identifies degraded retrieval keys, and a differentiable surrogate of \ac{LAC} can be used to train a better retrieval key.
Overall, our results suggest that retrieval baselines should be reported alongside \ac{VLA} performance to distinguish benchmark memorization from generalization.

These conclusions remain limited to \ac{LIBERO} and related benchmarks.
Their relatively dense demonstration coverage may favor retrieval-based methods.
LIBERO-Plus perturbations also largely preserve the underlying task and solution trajectories, so our robustness results mainly concern changes in observations and instructions, rather than situations that require new trajectories.
In addition, we evaluated the public checkpoints of each \ac{VLA} as released, and some may perform differently from their published results.
More fundamentally, retrieval cannot extrapolate beyond the support of its memory, and the strong performance observed on these benchmarks should not be interpreted as evidence that \ac{kNN} policies can generalize broadly across robotic problems.

Future work should extend this analysis to other robot-learning benchmarks, especially settings with sparser demonstrations, novel object placements~\citep{zhou2025liberopro,garcia2025gembench}, and tasks requiring behavior outside the demonstrated support.
Such evaluations would help determine how much of the observed behavior is specific to \ac{LIBERO} and how much reflects more general properties of current \ac{VLA} policies.

\clearpage
\acknowledgments{The authors thank the \href{https://github.com/huggingface/lerobot}{LeRobot} community for developing and maintaining the open-source framework used to evaluate the \ac{VLA} policies, and the authors of the evaluated models and benchmarks for releasing their checkpoints and data. This work was supported by \href{https://sereact.ai}{Sereact GmbH}.}
\bibliography{refs}

\clearpage
\appendix

\crefalias{section}{appendix}
\crefalias{subsection}{subappendix}

\section{Additional details of the \ac{kNN} policy}\label{app:knn}

This appendix provides the implementation details and ablations of the \ac{kNN} policy introduced in \cref{sec:methodology}.
We use the same notation throughout.

\subsection{Distance normalization}

The modality-specific scaling factors in \cref{eq:retrieval-distance} are estimated once when constructing the retrieval memory.
Under a median normalization, the visual scale for camera $c$ is
\begin{equation}
    s_c = \median_{\substack{i,j:\tau(i)=\tau(j)\\i\neq j}} \dist{\vect{z}_c^i-\vect{z}_c^j}_2^2.
\end{equation}

Similarly, the state and temporal scales are defined as
\begin{equation}
    s_x = \median_{\substack{i,j:\tau(i)=\tau(j)\\ i\neq j}} \frac{1}{n_x} \dist{\tilde{\vect{x}}^i-\tilde{\vect{x}}^j}_2^2,
    \label{eq:scale-state}
\end{equation}
and
\begin{equation}
    s_t
    =
    \median_{\substack{i,j:\tau(i)=\tau(j)\\ i\neq j}}
    (t_i-t_j)^2.
    \label{eq:scale-time}
\end{equation}

For language, the scale is computed over pairs of distinct task
instructions:
\begin{equation}
    s_\ell
    =
    \median_{\tau\neq\tau'}
    \dist{\vect{u}_{\ell_\tau} - \vect{u}_{\ell_{\tau'}}}_2^2.
    \label{eq:scale-language}
\end{equation}

The gripper term is not normalized by a modality-specific scale.

\subsection{History matching and trajectory consistency}

The full retrieval policy can incorporate recent observation history to disambiguate states that look similar but occur at different stages of a trajectory.

Let $q'$ denote a previous query observation separated from $q$ by $\Delta$ time steps.
For a candidate memory entry $i$, let $i^{-\Delta}$ denote the entry $\Delta$ steps earlier in the same demonstration.
We define
\begin{equation}
    d_1(q,i) = d_0(q,i) + w_h \bar{d}_0(q',i^{-\Delta}),
    \label{eq:history-distance}
\end{equation}
where $\bar d_0$ is the base retrieval distance without the language term.

We optionally favor candidates that continue the trajectory selected at the previous control step.
Let $(m^\star,t^\star)$ denote the expected continuation of the previous match.
The final distance is
\begin{equation}
    d(q,i)
    =
    d_1(q,i)
    \left(
    1 - w_s
    \mat{1} \left[m_i=m^\star, |t_i-t^\star|\leq W\right]
    \right),
    \label{eq:trajectory-consistency}
\end{equation}
where $W$ is the temporal consistency window.

\subsection{Neighbor weighting}

In the implementation, the softmax temperature is scaled by the mean distance of the retrieved neighborhood.
Let
\begin{equation}
    \bar{d} = \frac{1}{k} \sum_{j\in\mathcal{N}_k(q)}
    d(q,j).
    \label{eq:mean-neighbour-distance}
\end{equation}
The neighbor weights are then
\begin{equation}
    \alpha_i =
    \frac{
        \exp\left(-d(q,i)/(T\bar d)\right)
    }{
        \sum_{j\in\mathcal{N}_k(q)}
        \exp\left(-d(q,j)/(T\bar d)\right)
    }.
    \label{eq:exact-neighbour-weights}
\end{equation}

This local rescaling makes the weighting less sensitive to the absolute scale of distances across queries.

\subsection{Hyperparameter tuning}\label{app:tuning}

\begin{figure}[htbp]
    \centering
    \begin{subfigure}[t]{0.48\linewidth}
        \centering
        \includegraphics[width=\linewidth]{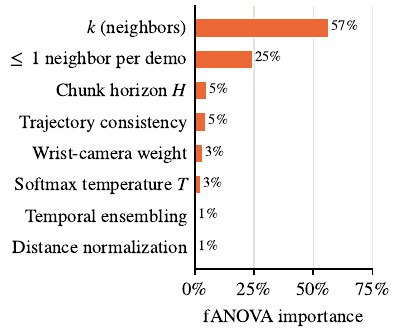}
        \caption{Hyperparameter importance estimated with fANOVA~\citep{hutter2014fanova}.}
        \label{subfig:optuna-importance}
    \end{subfigure}
    \hfill
    \begin{subfigure}[t]{0.48\linewidth}
        \centering
        \includegraphics[width=\linewidth]{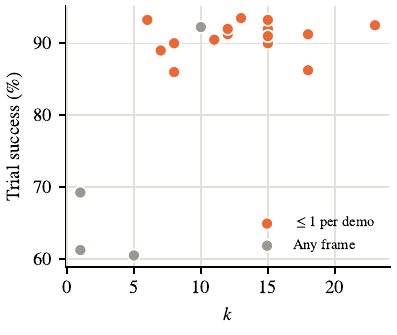}
        \caption{Success rate as a function of $k$.}
        \label{subfig:optuna-k}
    \end{subfigure}
    \caption{Hyperparameter search for the \ac{kNN} policy with Optuna, over \num{\OptunaN} completed trials of 400 episodes.
        (\subref{subfig:optuna-importance}) Hyperparameter importance estimated with fANOVA~\citep{hutter2014fanova}.
        (\subref{subfig:optuna-k}) Success rate as a function of the number of neighbors $k$ and of the uniqueness constraint of \cref{eq:unique-demo}.
    }
    \label{fig:optuna}
\end{figure}

We tune the retrieval hyperparameters with Optuna~\citep{akiba2019optuna}, using success rate as the optimization objective.
As \ac{LIBERO} provides no separate validation split, each trial is evaluated on the initial states 40--49 of every task (400 episodes), which are also part of the official evaluation protocol.
On the initial states 0--39, never used during tuning, the \emph{multimodal} variant reaches \SI{\KnnUnseenStates}{\percent}, against \SI{\KnnAvg}{\percent} on all 50 states, so tuning on these states does not noticeably inflate the reported results.
\Cref{subfig:optuna-importance} shows that performance is driven primarily by the number of neighbors $k$ and by the retrieval restriction of at most one neighbor per demonstration as defined in \cref{eq:unique-demo}, accounting for \SI{57}{\percent} and \SI{25}{\percent} of the fANOVA importance~\citep{hutter2014fanova}, respectively.
The remaining hyperparameters each account for about \SI{5}{\percent} or less.
\Cref{subfig:optuna-k} shows that using the per-demonstration constraint and a number of neighbors $k$ greater than 5 are indicative of high success rates.

\subsection{Ablation study}\label{app:ablations}

\Cref{tab:ablations} evaluates each component independently by changing one element of the \emph{multimodal} variant at a time.
The largest drops come from replacing the ResNet-18 features with raw $16\times16$ pixels (\SI{-34.5}{\pp}), ignoring the language prompt (\SI{-27.1}{\pp}), and reducing the demonstration memory to \SI{10}{\percent} (\SI{-14.3}{\pp}).
In contrast, mode seeking, temporal ensembling, history matching, proprioception, and the time-step feature each change average success by at most \SI{1}{\pp}, within statistical noise.
For $k$ between 4 and 32, success remains between \SI{89.4}{\percent} and \SI{92.7}{\percent}, whereas replaying the nearest demonstration ($k=1$) lowers it to \SI{78.6}{\percent}.

\begin{table}[htbp]
    \centering\footnotesize\setlength{\tabcolsep}{4pt}
    \caption{Ablations of the \emph{multimodal} \ac{kNN} variant under the standard \ac{LIBERO} evaluation protocol (50 rollouts per task, \num{2000} episodes in total).
        $\Delta$ denotes the change in average success relative to the multimodal variant.
        Raw-pixel retrieval fails on LIBERO-Object, where the ten tasks share a scene containing several small objects.
    }\label{tab:ablations}
    \begin{tabularx}{\linewidth}{@{}X*{6}{r}@{}}
\toprule
Ablation & avg & Spatial & Object & Goal & Long & $\Delta$ \\
\midrule
None (multimodal variant) & 91.7 & 88.0 & 99.8 & 97.6 & 81.4 & \\
Prompt ignored & 64.6 & 77.6 & 99.6 & 8.8 & 72.4 & −27.1 \\
Raw 16$\times$16 pixels & 57.2 & 79.0 & 0.0 & 91.0 & 58.6 & −34.5 \\
10\% of demos & 77.4 & 83.8 & 87.0 & 90.2 & 48.6 & −14.3 \\
25\% of demos & 82.6 & 83.6 & 97.8 & 92.6 & 56.4 & −9.1 \\
50\% of demos & 88.5 & 86.4 & 99.6 & 95.4 & 72.6 & −3.2 \\
$k = 1$ (replay nearest demo) & 78.6 & 77.2 & 90.6 & 86.2 & 60.4 & −13.1 \\
DINOv2-S~\citep{oquab2024dinov2} instead of ResNet-18 & 83.4 & 79.8 & 97.6 & 95.0 & 61.0 & −8.3 \\
Agent-view camera only & 82.1 & 78.0 & 92.8 & 93.6 & 63.8 & −9.6 \\
Wrist camera only & 87.5 & 87.0 & 98.8 & 92.8 & 71.2 & −4.2 \\
Plain weighted mean & 90.7 & 88.4 & 100.0 & 96.6 & 77.8 & −1.0 \\
No temporal ensembling & 91.1 & 92.0 & 98.0 & 94.2 & 80.2 & −0.6 \\
No history matching & 91.5 & 89.6 & 98.8 & 97.6 & 79.8 & −0.2 \\
No proprioceptive state & 91.4 & 87.6 & 99.6 & 96.8 & 81.4 & −0.3 \\
No time-step feature & 91.3 & 88.2 & 99.8 & 97.4 & 79.8 & −0.4 \\
Hard task filter & 91.9 & 88.4 & 100.0 & 97.4 & 81.8 & +0.2 \\
\midrule
$k = 2$ & 85.2 & 83.8 & 94.6 & 92.2 & 70.2 & −6.5 \\
$k = 4$ & 89.4 & 86.2 & 98.8 & 93.8 & 78.6 & −2.3 \\
$k = 8$ & 89.8 & 85.4 & 99.6 & 96.6 & 77.6 & −1.9 \\
$k = 24$ & 92.7 & 89.2 & 99.8 & 97.2 & 84.4 & +1.0 \\
$k = 32$ & 91.7 & 88.8 & 99.8 & 97.4 & 80.8 & +0.0 \\
\bottomrule
\end{tabularx}

\end{table}

\section{\ac{VLA} evaluation protocol}\label{app:vlas}

We evaluated nine publicly available \ac{VLA} checkpoints as released.
\Cref{tab:checkpoints} lists the Hugging Face repository and revision of each evaluated checkpoint.
For $\pi_{0.5}$, $\pi_0$, and $\pi_0$-FAST, we use the LIBERO fine-tuned checkpoints from \href{https://github.com/huggingface/lerobot}{LeRobot}~\citep{LeRobot2026}.
X-VLA uses absolute end-effector control and its native observation convention.
VLA-JEPA executes 7 actions per re-plan, while Evo-1 executes 14.
For Evo-1, we use \SI{256}{\pixel} observations on both LIBERO and LIBERO-Plus, consistent with the image resolution assumed by the LIBERO-Plus perturbations.
MolmoAct2 outputs continuous actions.
For GR00T N1.7, we use one checkpoint per LIBERO suite and execute 8 of its 16 predicted actions before re-planning.
Some reproduced success rates, notably for $\pi_0$ and SmolVLA, are below those reported by their authors, which may reflect differences between the public checkpoints and those used in the original publications.

\begin{table}[t]
    \centering
    \caption{Evaluated \ac{VLA} checkpoints, hosted on the Hugging Face Hub. Revisions are abbreviated commit hashes. GR00T N1.7 uses one checkpoint per LIBERO suite.}\label{tab:checkpoints}
    \begin{tabularx}{\linewidth}{@{}l>{\ttfamily\raggedright\arraybackslash}X>{\ttfamily}l@{}}
        \toprule
        Model        & \normalfont Repository                     & \normalfont Revision \\
        \midrule
        $\pi_{0.5}$  & lerobot/pi05\_libero\_finetuned\_v044      & 8e174154             \\
        MolmoAct2    & allenai/MolmoAct2-LIBERO-LeRobot           & f0c5a956             \\
        GR00T N1.7   & nvidia/gr00t17-lerobot-libero\_spatial-640 & 32a6ec78             \\
                     & nvidia/gr00t17-lerobot-libero\_object-640  & 1499db35             \\
                     & nvidia/gr00t17-lerobot-libero\_goal-640    & 436c57c0             \\
                     & nvidia/gr00t17-lerobot-libero\_10-640      & 5ee08ab0             \\
        VLA-JEPA     & lerobot/VLA-JEPA-LIBERO                    & 735d9f69             \\
        Evo-1        & zuoxingdong/evo1\_libero                   & 515921f4             \\
        X-VLA        & lerobot/xvla-libero                        & 12e8783e             \\
        $\pi_0$      & lerobot/pi0\_libero\_finetuned\_v044       & 45dcc8fc             \\
        SmolVLA      & lerobot/smolvla\_libero                    & 31d453f7             \\
        $\pi_0$-FAST & lerobot/pi0fast-libero                     & 840f4b50             \\
        \bottomrule
    \end{tabularx}
\end{table}

Before closed-loop evaluation, we performed a simple offline sanity check on each checkpoint.
On demonstration frames, the predicted motion must be closer to the corresponding demonstrated motion than to a shuffled pairing, with an error below half of the shuffled baseline.
This check is used only to detect clearly invalid model or preprocessing configurations before evaluation.

\section{Full results}\label{app:full-results}

\begin{figure}[htbp]
    \centering
    \includegraphics[width=\linewidth]{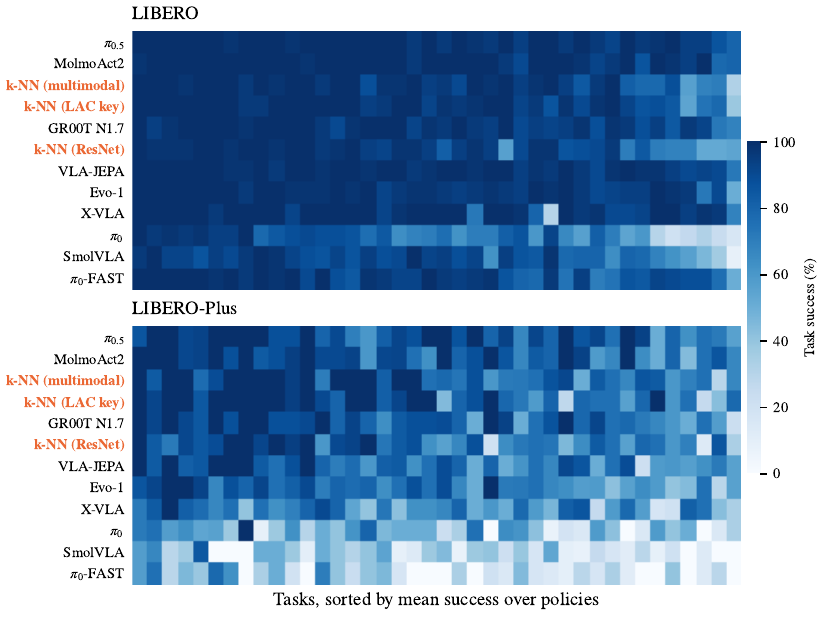}
    \caption{Success of every policy on each of the 40 tasks, tasks sorted from easiest to hardest (mean over policies).}\label{fig:success-tasks}
\end{figure}

\Cref{fig:success-tasks} shows the success rate of every policy on each of the 40 tasks of \ac{LIBERO} and LIBERO-Plus, sorted from easiest to hardest.
For both benchmarks, $\pi_{0.5}$ and MolmoAct2 are the best policies.

\begin{figure}[htbp]
    \centering
    \includegraphics[width=\linewidth]{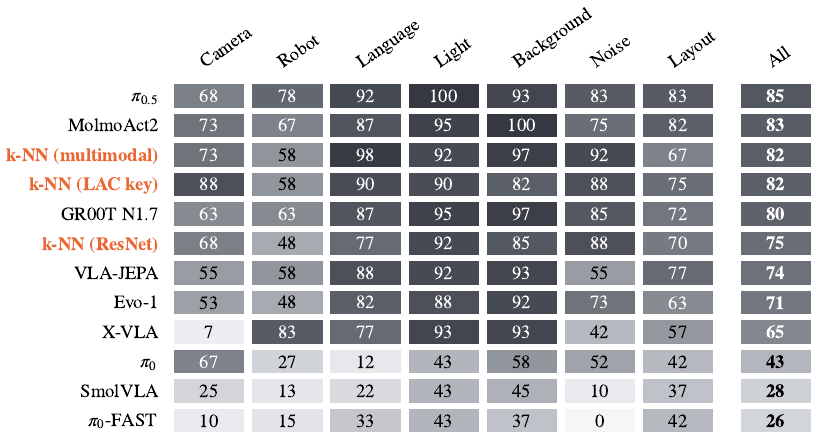}
    \caption{Success rate \parunit{\percent} of each policy on LIBERO-Plus tasks, per type of perturbation.}\label{fig:success-perturbations}
\end{figure}

\Cref{fig:success-perturbations} shows the success rate of each policy on LIBERO-Plus, per perturbation category.
As in \cref{fig:success-tasks}, $\pi_{0.5}$ and MolmoAct2 are the most successful on average, and most policies are resilient to perturbed lighting.
Camera perturbations reduce the success of all policies, except for the \emph{LAC key} variant, which reaches \SI{88}{\percent} against at most \SI{73}{\percent} for the others.

\section{Embedding space visualization}\label{app:maps}

\begin{figure}[htbp]
    \centering
    \includegraphics[width=\linewidth]{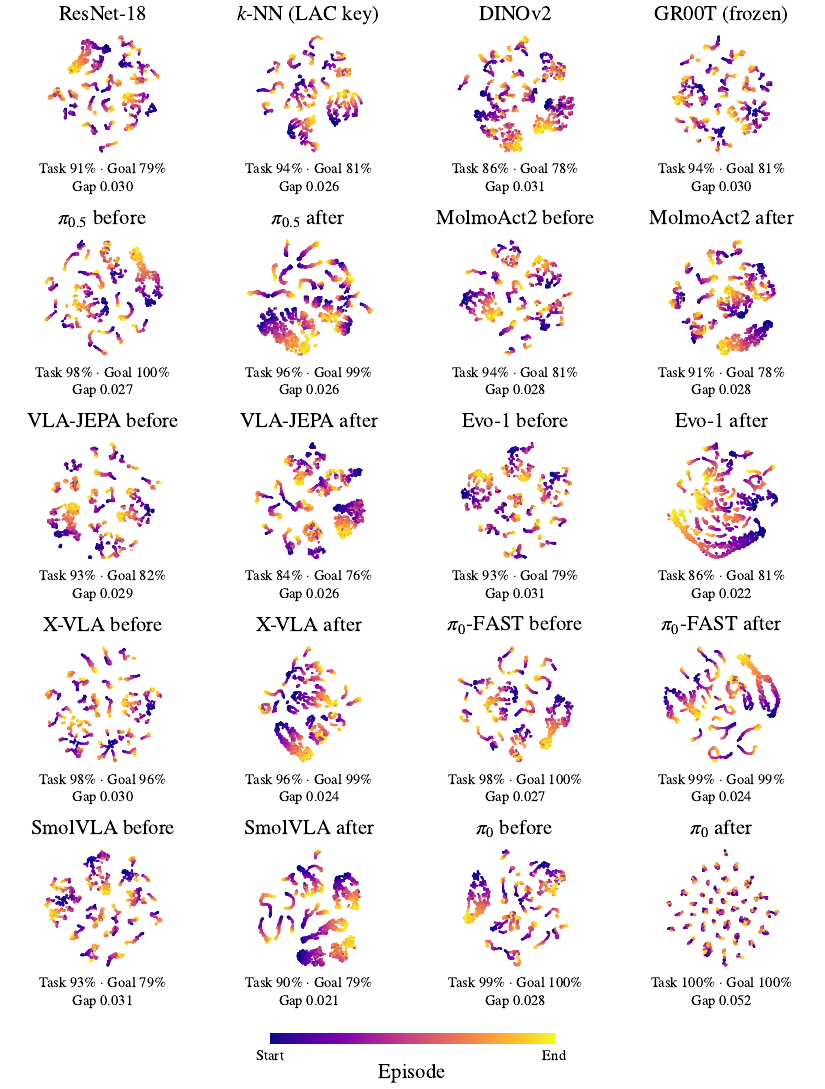}
    \caption{t-SNE visualization of each retrieval space, colored by progress through the episode.
        Below each map: Task, the fraction of nearest neighbors from another demonstration that belong to the same task; Goal, the same fraction within LIBERO-Goal; Gap, the median progress difference to the nearest frame from another demonstration of the same task.
        GR00T freezes its \ac{VLM}, so its representation is unchanged by policy fine-tuning.
    }
    \label{fig:maps-tsne}
\end{figure}

\begin{figure}[htbp]
    \centering
    \includegraphics[width=\linewidth]{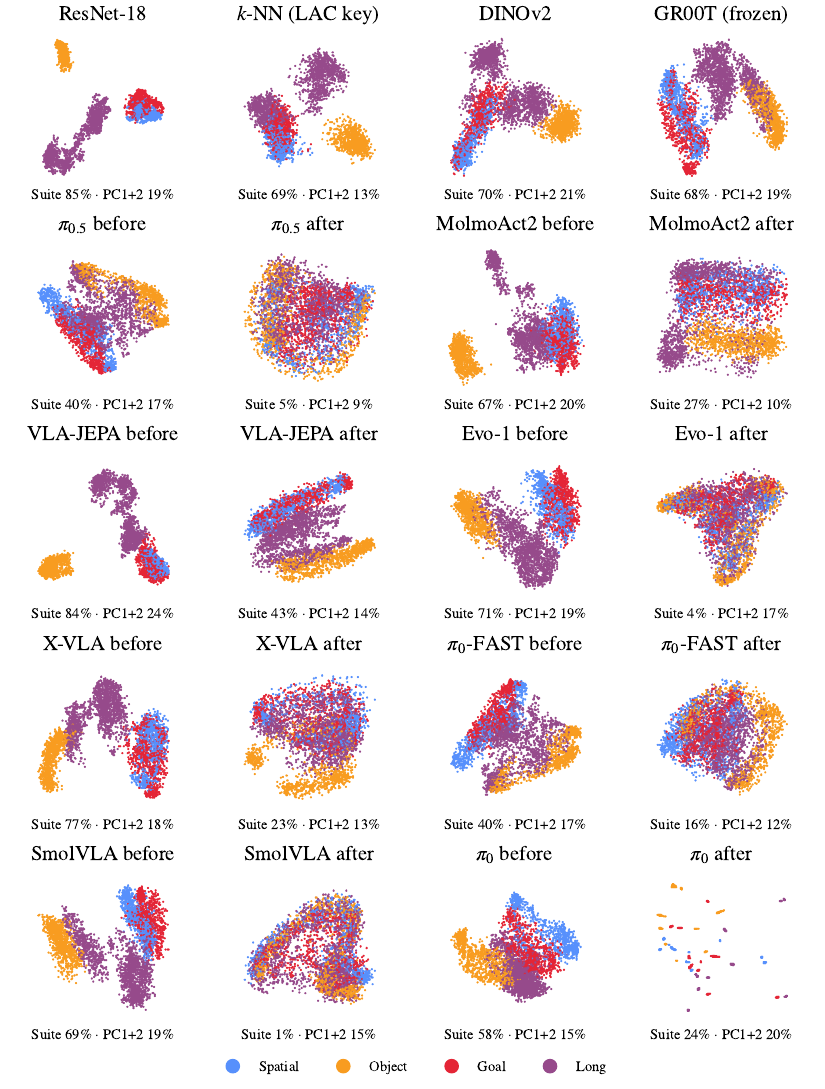}
    \caption{
        First two principal components of each retrieval key, colored by LIBERO suite.
        Below each map: Suite, the fraction of the variance of the two components explained by the suite; PC1+2, the fraction of the total variance captured by the two components.
    }
    \label{fig:maps-pca}
\end{figure}

We visualize the embedding spaces of all keys reported in \cref{tab:backbones}, together with the generic visual keys, on the same set of \MapN{} demonstration frames as in \cref{fig:maps}.
We sample six demonstrations per task and retain every tenth frame.
Each key is computed exactly as for the \emph{ResNet} variant \pcref{sec:implementation}: features are standardized using memory statistics, unit-normalized independently for each camera, and then concatenated across cameras.
\Cref{fig:maps-tsne} shows \ac{t-SNE} embeddings computed from the first 50 principal components with perplexity 30, while \cref{fig:maps-pca} shows the first two principal components.

Before \ac{LIBERO} fine-tuning, the embedding spaces are strongly structured by suite, and therefore by scene.
The suite explains \num{\MapSuiteBeforeMin}-\SI{\MapSuiteBeforeMax}{\percent} of the variance in the first two principal components across \ac{VLM} keys, compared with \SI{\MapSuiteResnet}{\percent} for an ImageNet-pretrained ResNet-18.
After fine-tuning, this fraction decreases to \num{\MapSuiteAfterMin}-\SI{\MapSuiteAfterMax}{\percent}, with a decrease for \num{\MapSuiteDown} of the \num{\MapPhaseN} fine-tuned \acp{VLM}.
These two components capture \num{\MapPcaVarMin}-\SI{\MapPcaVarMax}{\percent} of the total variance of each key, so the suite dominates the main directions of variation but not the full representation.

For every key, \num{\MapTaskMin}-\SI{\MapTaskMax}{\percent} of the nearest neighbors from other demonstrations belong to the same task.
Within LIBERO-Goal, where the ten tasks share one scene, the pattern differs.
Only representations whose image tokens can attend to the instruction clearly separate its ten tasks.
This includes the bidirectional prefix attention used by PaliGemma~\citep{beyer2024paligemma} in $\pi_0$, $\pi_{0.5}$, and $\pi_0$-FAST, as well as the Florence-2 encoder~\citep{Xiao2024} used by X-VLA.
These representations yield \num{\MapGoalPromptMin}-\SI{100}{\percent} same-task nearest neighbors.
In contrast, \acp{VLM} with causal attention, where image tokens precede the instruction and therefore cannot attend to it, achieve only \num{\MapGoalCausalMin}-\SI{\MapGoalCausalMax}{\percent}, while ResNet-18 achieves \SI{\MapGoalResnet}{\percent}.
The median progress gap is discussed in \cref{sec:representation-analysis}.

\section{\ac{LAC} and closed-loop retrieval}\label{app:lac}

\begin{figure}[htbp]
    \centering
    \includegraphics[width=\linewidth]{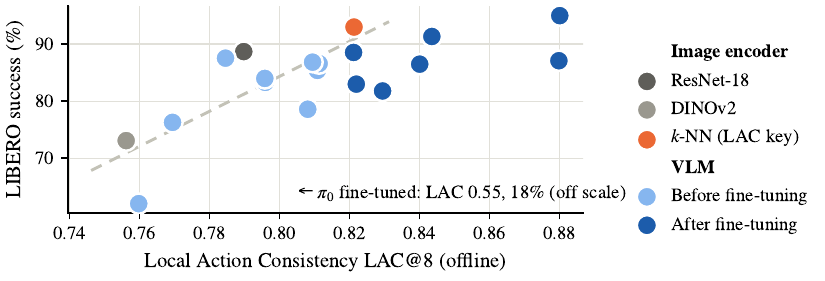}
    \caption{Offline \ac{LAC} versus closed-loop retrieval success.
        Each point is a visual representation used as the key of the \emph{ResNet} \ac{kNN} variant.
        The x-axis reports $\mathrm{LAC@}8$ on held-out demonstrations, and the y-axis the \ac{LIBERO} success rate.
        The dashed line is fitted on all keys except the fine-tuned \ac{VLM} backbones, which were trained on the held-out demonstrations.
    }\label{fig:neighbors}
\end{figure}

For each visual representation, we compute \ac{LAC} offline on held-out demonstration queries and independently evaluate the \emph{ResNet} \ac{kNN} variant using the same representation as its retrieval key.
\Cref{fig:neighbors} compares these quantities across generic visual encoders and \ac{VLM} backbones before and after \ac{LIBERO} fine-tuning.
Higher \ac{LAC} generally corresponds to higher closed-loop retrieval success, supporting its use as an offline measure of retrieval-key quality.

\end{document}